\documentclass[runningheads]{llncs}
\usepackage{amssymb}
\usepackage{amsmath}
\usepackage{url}
\usepackage[T1]{fontenc}
\usepackage{graphicx}
\usepackage{comment}
\usepackage{tikz}
\usetikzlibrary{arrows.meta,positioning}
\newcommand{\RF}{RF}
\newcommand{\LR}{LR}
\newcommand{\ET}{ExtraTrees}
\newcommand{\GBDT}{GBDT}
\newcommand{\AdaBoost}{AdaBoost}
\newcommand{\SVM}{SVM}
\newcommand{\SVMLIN}{SVM-Lin}
\newcommand{\SVMRBF}{SVM-RBF}
\newcommand{\KNN}{kNN}
\newcommand{\NB}{NB}
\newcommand{\MLP}{MLP}

\makeatletter
\renewcommand\section{\@startsection{section}{1}{\z@}%
	{0.3ex \@plus 0ex \@minus 0ex}%
	{0.3ex \@plus 0.1ex}%
	{\normalfont\large\bfseries}}
\renewcommand\subsection{\@startsection{subsection}{2}{\z@}%
	{0.2ex \@plus 0ex \@minus 0ex}%
	{0.2ex \@plus 0.1ex}%
	{\normalfont\normalsize\bfseries}}
\renewcommand\subsubsection{\@startsection{subsubsection}{3}{\z@}%
	{0.2ex \@plus 0ex \@minus 0ex}%
	{0.2ex \@plus 0.1ex}%
	{\normalfont\normalsize\itshape}}
\makeatother

\begin{document}
	\title{When Can We Trust Early Warnings? Leakage-Excluded Early Outcome Prediction from LMS Interaction Logs\vspace{-0.7cm}}
	%
	%
	%
	\authorrunning{F. Author et al.}
	\titlerunning{Leakage-Excluded Early Outcome Prediction from LMS Interaction Logs\vspace{-0.8cm}}
	\author{Ng\d{o}c Luy\d{\^{e}}n L\^e\inst{1,2} \and
		Marie-H\'el\`ene Abel\inst{2} \and
		Bertrand Laforge\inst{3}}
	\authorrunning{NL Le et al.}
	%
	\institute{Gamaizer, 93340 Le Raincy, France \and
		Universit\'{e} de technologie de Compi\`egne, CNRS, Heudiasyc (Heuristics and Diagnosis of Complex Systems), CS 60319 - 60203 Compi\`egne Cedex, France
		\and
		Sorbonne Universit\'{e}, CNRS UMR 7585, LPMHE (Laboratoire de Physique Nucl\'{e}aire et des Hautes \'{E}nergies), 75252 Paris cedex 05, France\\
	}
	\maketitle              
	\vspace{-0.8cm}

\begin{abstract}
	Early-warning models built from Learning Management System (LMS) logs aim to predict end-of-course outcomes early enough to enable timely learner support. However, reported ``early'' performance is often inflated by \emph{temporal leakage}. This occurs when the pipeline uses information that would not yet be available at the time of prediction. We formalize cutoff-based early outcome prediction under a temporal availability constraint and introduce \textbf{LEAP} (Leakage-Excluded Early-Availability Protocol), which enforces \emph{cutoff-first} truncation prior to joins and aggregation and audits feature provenance to prevent post-cutoff evidence from entering the benchmark. We instantiate LEAP on the public Open University Learning Analytics Dataset (OULAD) as a multi-step protocol for leakage-controlled evaluation across weekly cutoffs. Using several standard learning methods, we evaluate performance using ROC-AUC, PR-AUC, Brier score, and F1@0.5. Results show improving performance as the observation window expands, with a marked gain around week~3; Random Forest performs best at the earliest cutoffs, while Gradient Boosting dominates thereafter. Leakage ablations further show that temporal violations, especially through assessment information, can inflate apparent ``early'' performance.
	
	\vspace{-0.3cm}	
\end{abstract}
	\vspace{-0.6cm}	
\section{Introduction}\label{sec:introduction}
Learning Management Systems (LMS) have become central infrastructures for online and blended education, generating large volumes of time-stamped interaction data that reflect how learners engage with resources, activities, and assessments. These behavioral traces have motivated extensive work in Learning Analytics, Educational Data Mining, and Intelligent Tutoring Systems on early warning and performance prediction, with the objective of identifying learners in difficulty early enough to enable timely support~\cite{chaka2021educational}. Prior studies suggest that meaningful predictive signals can emerge within the first weeks of a course and that models trained on log-derived features can provide useful discrimination early on, improving as additional evidence accumulates over time~\cite{bernacki2020predicting,santos2023accurate}.

This issue of temporal validity is especially consequential in educational early-warning settings. Unlike many standard predictive tasks, the purpose of an early-warning model is not only to predict accurately, but to do so early enough to support intervention. LMS data also combine heterogeneous sources -- behavioral traces, pedagogical activities, and assessment records -- whose observability evolves differently over the course timeline. As a result, seemingly minor preprocessing decisions can change whether a benchmark truly reflects an actionable early-warning scenario or an unrealistically optimistic retrospective prediction.

A key challenge, however, is ensuring that reported results correspond to the intended early-warning setting. In practice, ``early'' performance can be inflated by temporal leakage, which occurs when the modeling pipeline incorporates information that would not be available at the prediction time~\cite{kaufman2012leakage}. Leakage may arise when time truncation is applied after aggregation, when post-cutoff assessments are mixed into early features, or when evaluation choices allow future information to influence training and testing. Such violations can yield overly optimistic estimates that do not transfer to real-time deployment and undermine the trustworthiness of early warning systems in educational practice~\cite{10.1093/jamia/ocad178,esbenshade2024non}.


This paper studies early outcome prediction under strict temporal validity. We introduce \textbf{LEAP}, a protocol for leakage-controlled benchmarking of cutoff-based early prediction from LMS logs. LEAP enforces time truncation before joins, aggregation, and feature construction, and verifies feature provenance through explicit timestamp-based checks. We instantiate LEAP on OULAD~\cite{kuzilek2017open} and evaluate standard benchmark models across weekly cutoffs, together with controlled leakage ablations.
More specifically, this paper makes three contributions: (i) it defines LEAP as a protocol for early prediction under explicit temporal-availability constraints; (ii) it shows how predictive performance evolves across weekly cutoffs under strict leakage control; and (iii) it quantifies how much apparent ``early'' performance can be inflated when post-cutoff information enters the pipeline.

The central question of this paper is when early warnings can be considered trustworthy under realistic temporal constraints. To address this question, we consider four research questions: (\textit{RQ1}) how predictive performance evolves as the observation window expands under strict temporal availability? (\textit{RQ2}) which standard model families perform most reliably at different cutoffs? (\textit{RQ3}) how strongly temporal leakage inflates reported ``early'' performance and which sources drive this inflation? and (\textit{RQ4}) how the predictive signal shifts over time from behavioral engagement toward assessment-related evidence?

The remainder of the paper is organized as follows. Section~\ref{sec:related_work} reviews related work and highlights temporal-validity and reproducibility risks. Section~\ref{sec:methodology} introduces the cutoff-based setting and LEAP. Section~\ref{sec:experiments} presents the experimental setup, and Section~\ref{sec:results} reports results across cutoffs, including leakage ablations and signal analysis. Section~\ref{sec:conclusion} concludes with implications and future work.
	
\section{Related Work}
\label{sec:related_work}

Early warning and early outcome prediction have extensively used LMS interaction logs to anticipate end-of-course success or withdrawal. Prior work shows that interpretable activity indicators such as resource views, forum participation, and assessment engagement can support effective prediction and intervention \cite{macfadyen2010mining,hu2014developing,bernacki2020predicting,akccapinar2019using}. Kuzilek et al.\ released OULAD as a public benchmark dataset to support reproducible research \cite{kuzilek2017open}. More recent studies evaluate multiple early cutoffs to characterize the timing--accuracy trade-off and examine portability across courses and course runs \cite{santos2023accurate,conijn2016predicting}.

A continuing challenge is ensuring comparability and deployment realism, because studies differ in how observation windows are defined and how temporal validity is enforced. Temporal leakage can arise not only from explicit post-cutoff features, but also from pipeline choices such as aggregating before truncation, normalizing with full-course statistics, or joining sources in ways that encode future events~\cite{kaufman2012leakage}. These violations can inflate reported early performance~\cite{10.1093/jamia/ocad178,esbenshade2024non}, reduce reproducibility~\cite{kapoor2023leakage}, and undermine trustworthy deployment~\cite{tiggeloven2025role}. Similar risks also arise at the evaluation stage, for example when future-overlap effects are not controlled across cohorts or course runs~\cite{esbenshade2024non}. Yet temporal validity is often described only in broad terms, without specifying where truncation occurs, how preprocessing is isolated, or how post-cutoff evidence is verified to be absent.

Although prior work shows that LMS interactions support early prediction and that multi-cutoff evaluation is informative, a methodological gap remains: the availability constraint is rarely made explicit at each cutoff, and leakage control is often treated as an assumption rather than an enforced requirement. LEAP is intended as a concrete protocol-level response to this gap.

\section{Methodology: Leakage-Excluded Early Prediction}
\label{sec:methodology}

We address the problem of \emph{early} prediction of learners’ final outcomes from partially observed interaction traces in an LMS~\cite{santos2023accurate,bernacki2020predicting}. The central methodological requirement is \emph{temporal validity}: for any early cutoff time $t$, defined as the end of the observation window up to which LMS records are observed, the predictive model must use only information that would be available up to time $t$~\cite{10.1093/jamia/ocad178}. This constraint ensures that performance estimates reflect realistic early warning conditions and are not inflated by inadvertent access to future information~\cite{esbenshade2024non}.

To operationalize this requirement, we introduce \textbf{LEAP} (\emph{Leakage-Excluded Early-Availability Protocol}), which enforces strict time truncation prior to feature construction and includes explicit checks preventing information beyond the observation window ending at time $t$ from entering the modeling pipeline. In this paper, we use the term \emph{protocol} to denote an explicit set of preprocessing and validation rules that must hold at each cutoff before any model is trained or evaluated.

A simple way to think about LEAP is the following. Suppose we want to issue a prediction at day 14. A valid early-warning pipeline must first remove every interaction, submission, or assessment-related record occurring after day 14 from all source tables, and only then compute aggregates such as total clicks, number of active days, or number of submissions. A leaky pipeline may appear to use a day-14 cutoff while still encoding future information indirectly -- for example by aggregating over the full course before truncation, by joining assessment tables that include later submissions, or by computing preprocessing statistics from the full dataset. LEAP is designed to rule out such cases explicitly.

\subsection{Task Formulation}
\label{sec:task}
Let $\mathcal{I}$ denote the set of prediction instances. Each instance $i \in \mathcal{I}$ corresponds to a learner within a particular course run (e.g., specified by module, presentation, and student identifier). The target variable is a binary end-of-course outcome $y_i \in \{0,1\}$, defined after the course has concluded.

For each instance $i$, let $\mathcal{R}_i$ be the multiset of time-stamped LMS records\footnote{We use the term record to denote any time-stamped LMS entry, including interaction events and assessment logs.}, including both behavioral interactions and assessment-related records. Each record $r \in \mathcal{R}_i$ is associated with a timestamp $\tau(r) \in \mathbb{R}$ (e.g., day index relative to the course start).

Given a cutoff time $t$, our goal is to learn a probabilistic predictor

\vspace{-0.2cm}
\begin{equation}
	\hat{p}_i^{(t)} \;=\; f^{(t)}\!\left(x_i^{(t)}\right)\in[0,1],
	\vspace{-0.1cm}
\end{equation}
where $x_i^{(t)}$ is an early representation computed exclusively from records observed up to time $t$.
In other words, the prediction at cutoff $t$ must be based only on evidence observable by time $t$, not on information reconstructed retrospectively after the course has progressed further.

\begin{figure}[h]
	\centering
	\vspace{-0.7cm}
	\includegraphics[width=0.80\linewidth]{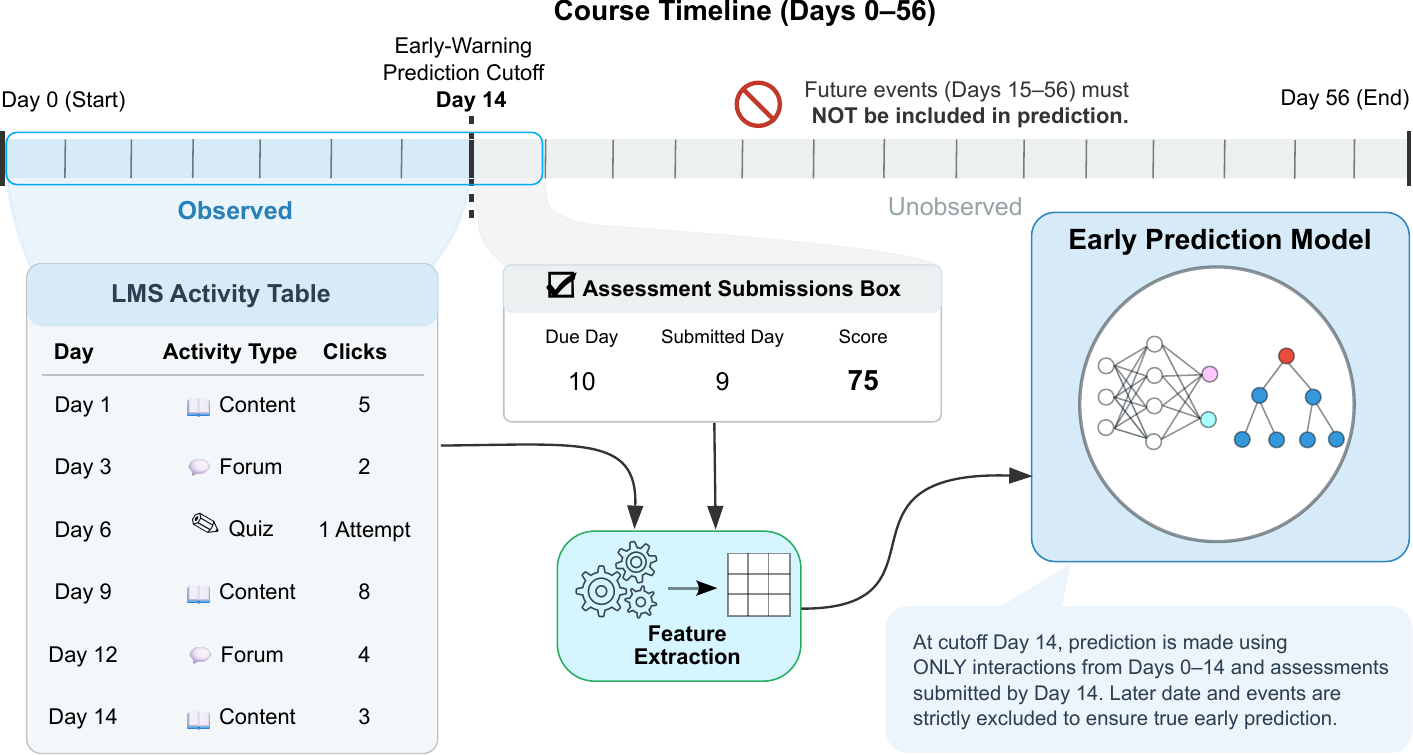}
	\vspace{-0.3cm}
	\caption{Example of an observation window ending at Day 14: only records with $\tau \le 14$ are observed; later records (Days 15–56) are excluded to prevent temporal leakage.}
	\label{fig:example}
	\vspace{-0.8cm}
	\end{figure}


In Figure~\ref{fig:example}, we illustrate the early prediction setting for a single learner--course instance. We assume the course begins at day $0$ and ends at day $56$, and we issue an early warning prediction at cutoff $t=14$ (end of week~2). At this time, only records with $\tau \le t$ are admissible; all later records must be excluded from feature construction, preprocessing, and model selection. These admissible records may reflect either practice activities, which can occur from the start of the course, or evaluative activities, such as graded quizzes and assignments, which become available only when scheduled and submitted. Consequently, early cutoffs may contain mainly engagement traces, while later cutoffs may also include assessment evidence.


\subsection{LEAP Constraint: Early Availability without Leakage}
\label{sec:constraint}
We formalize the information available at cutoff time $t$ and specify the condition under which early features are considered leakage-excluded. We define the set of records available by time $t$ as

\vspace{-0.4cm}
\begin{equation}
	\mathcal{R}_i^{(\le t)} \;=\; \{\, r \in \mathcal{R}_i \;:\; \tau(r)\le t \,\}.
\vspace{-0.1cm}
\end{equation}
An early representation $x_i^{(t)}$ is \emph{leakage-excluded} if it is a deterministic function of $\mathcal{R}_i^{(\le t)}$ only:
$
	x_i^{(t)} \;=\; \phi\!\left(\mathcal{R}_i^{(\le t)}\right),
$
for a fixed mapping $\phi$. In particular, no information derived from records with $\tau(r) > t$ may affect $x_i^{(t)}$.

In addition, LEAP constrains the order of preprocessing operations. Specifically, time truncation must be applied before any aggregation or transformation:

\vspace{-0.4cm}
\begin{equation}
	x_i^{(t)} \;=\; \phi\!\left(\mathcal{F}_t(\mathcal{R}_i)\right),
	\qquad
	\mathcal{F}_t(\mathcal{R}_i)=\mathcal{R}_i^{(\le t)}.
	\vspace{-0.1cm}
\end{equation}
Here, $\mathcal{F}_t$ denotes a deterministic time-truncation operator on records, whereas $f^{(t)}$ denotes the cutoff-specific predictive model operating on feature vectors. We next describe how this constraint is enforced within the pipeline. The key point is that leakage control is not only about \emph{which} records are allowed, but also about \emph{when} filtering occurs in the pipeline. Filtering after aggregation or after joining multiple sources may still leave future information encoded in the resulting features.

\subsection{Leakage Control}
\label{sec:guards}
LEAP enforces temporal validity via \emph{cutoff-first} preprocessing and a provenance check.
In practice, $\mathcal{R}_i$ is assembled from multiple timestamped data sources (e.g., interaction logs and assessment logs); we denote a generic source by $S$.
For each cutoff $t$, we first truncate each source to $S^{(\le t)}=\{r\in S:\tau(r)\le t\},$
and only then perform any join or aggregation (i.e., joins use truncated sources only, $S_1^{(\le t)} \Join S_2^{(\le t)}$).
Any learned preprocessing is fit on the training split only and applied to the test split within a pipeline.

To make the constraint verifiable, for each instance $i$ and feature group $g$ we record the \emph{provenance time}

\vspace{-0.6cm}
\begin{equation}
T_{i,g}^{(t)}=\max\{\tau(r): r \text{ is accessed to compute features\- in group } g\},	
\vspace{-0.4cm}
\end{equation}
and assert $T_{i,g}^{(t)}\le t$; violations halt the run.
When timestamps are not sufficient to guarantee observability, such fields are excluded in the strict setting unless an explicit availability assumption is stated.
In this sense, LEAP should be understood as an evaluation and preprocessing protocol rather than as a new predictive model. Its role is to make temporal-availability assumptions explicit, enforce truncation order, and provide checks against post-cutoff evidence. It does not by itself solve every possible form of deployment mismatch, but it aims to ensure that reported benchmark results correspond to the intended early-warning setting.

\subsection{Common Leakage Scenarios}
\label{sec:leakage_scenarios}

Temporal leakage in early-warning benchmarks does not arise only from obviously using records after cutoff $t$. In practice, leakage can also be introduced indirectly through the order of preprocessing operations, through joins across partially filtered sources, or through fields whose observability is unclear at prediction time. To make these risks more concrete, Table~\ref{tab:leakage_scenarios} summarizes several common leakage scenarios in LMS early prediction and indicates how LEAP guards against each of them.

\begin{table}[h]
	\centering
	\vspace{-0.7cm}
	\caption{Examples of temporal leakage scenarios in LMS early prediction and the corresponding LEAP safeguard.}
	\vspace{-0.2cm}
	\label{tab:leakage_scenarios}
	\begin{tabular}{| p{2.8cm} | p{5.0cm} | p{4.0cm} |}
		\hline
		\textbf{Scenario} & \textbf{Why it is leaky} & \textbf{LEAP safeguard} \\
		\hline
		Aggregation before truncation & Full-course activity is summarized before cutoff filtering, so future events may influence early features & Truncate each source at cutoff $t$ before any aggregation \\\hline
		Joins with unfiltered assessment tables & Later submissions or due-date-linked information may enter the joined table & Join only truncated sources $S_1^{(\le t)} \Join S_2^{(\le t)}$ \\\hline
		Preprocessing fit on full data & Scaling or transformation may encode information from the test split or later periods & Fit learned preprocessing on the training split only \\\hline
		Use of score fields with unclear observability & A score may not actually be visible at cutoff $t$ even if a submission exists & Exclude such fields unless an explicit observability assumption is stated \\
		\hline
	\end{tabular}\vspace{-0.7cm}
\end{table}

These examples clarify that leakage control is not limited to filtering rows by timestamp. It also requires controlling how features are constructed, how data sources are combined, and which variables can reasonably be considered observable at cutoff $t$.
\subsection{Early Representation}
\label{sec:repr}
The mapping $\phi(\cdot)$ produces a fixed-dimensional and interpretable vector $x_i^{(t)}$ that summarizes the learner’s activity within the observation window. The representation is designed to capture overall engagement, the diversity of interactions across activity types and learning resources, and the temporal distribution of activity over time. When available, the representation also incorporates early assessment evidence admissible at time $t$ (e.g., submissions and timing; scores only under an explicit availability assumption). Missingness is handled deterministically to preserve comparability across cutoffs.

\subsection{Evaluation Across Cutoffs}
\label{sec:earliness}
Let $\mathcal{T}=\{t_1,\dots,t_K\}$ denote a predefined set of early cutoffs. For each $t\in\mathcal{T}$, we construct a cutoff-specific dataset
$
	\mathcal{D}^{(t)} \;=\; \{(x_i^{(t)}, y_i)\}_{i\in\mathcal{I}^{(t)}},
$ where $\mathcal{I}^{(t)}$ contains instances for which an early representation is available at time $t$. To respect the information available at each cutoff, we train a separate predictor $f^{(t)}$ for each $t$. Repeating this process over $\mathcal{T}$ yields an earliness--performance profile that characterizes how predictive performance evolves as additional early evidence becomes available.

\begin{figure}[h]
	\centering
	\vspace{-0.2cm}
	\includegraphics[width=0.85\linewidth]{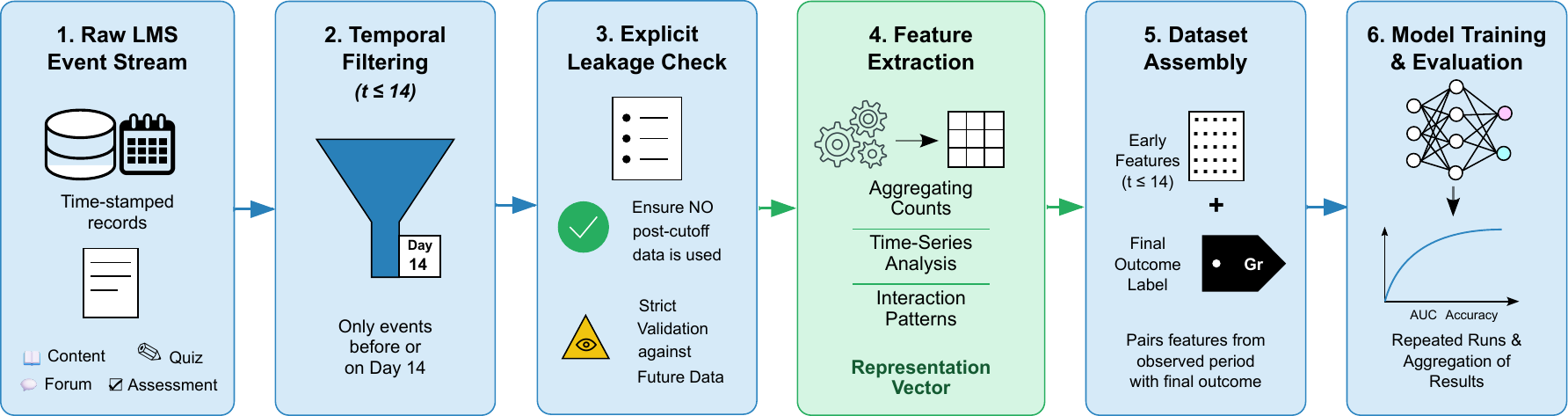}
	\vspace{-0.2cm}
	\caption{LEAP pipeline at cutoff $t$: time truncation precedes feature construction, leakage checks enforce temporal validity, and models are trained and evaluated per cutoff.}
	\label{fig:pipeline}
	\vspace{-0.7cm}
\end{figure}

\subsection{Pipeline Overview}
\label{sec:pipeline}
Figure~\ref{fig:pipeline} summarizes the LEAP pipeline at a given cutoff time $t$. For each learner--course instance, we start from the raw time-stamped LMS record stream $\mathcal{R}_i$ and apply a temporal filter to retain only records within the observation window, yielding $\mathcal{R}_i^{(\le t)}$. We then validate temporal integrity through leakage checks to ensure that no retained record occurs after $t$. The filtered record set is subsequently transformed into an early representation $x_i^{(t)}=\phi(\mathcal{R}_i^{(\le t)})$, which is paired with its end-of-course label to form the cutoff-specific dataset $\mathcal{D}^{(t)}$. This procedure is repeated for each $t \in \mathcal{T}$ to support benchmarking across multiple early observation windows. Building on this pipeline, the next section presents the experiments used to instantiate LEAP.

\section{Experiments}
\label{sec:experiments}
We describe the dataset, experimental setup, benchmark models, and evaluation metrics used in our study. The implementation is available\footnote{\url{https://github.com/lengocluyen/leap-early-prediction}}.

\subsection{Dataset}
\label{sec:dataset}

We conduct experiments on the Open University Learning Analytics Dataset (OULAD), which provides anonymized learner information, assessment data, and fine-grained Virtual Learning Environment (VLE) interaction logs~\cite{kuzilek2017open}. In OULAD, a course run is defined by the pair (\textit{code\_module}, \textit{code\_presentation}), and each prediction instance corresponds to one learner in one course run, indexed by (\textit{code\_module}, \textit{code\_presentation}, \textit{id\_student}).


We use \textit{studentInfo} to obtain end-of-course outcomes, \textit{studentVle} for interaction events, \textit{vle} for resource metadata, \textit{assessments} for assessment metadata, and \textit{studentAssessment} for submissions and scores. Interaction records are linked to resource types by joining \textit{studentVle.id\_site} with \textit{vle.id\_site}. When assessment evidence is used, submission records in \textit{studentAssessment} are joined with \textit{assessments} to access due-date metadata.

\begin{table}[h]
	\centering
	\vspace{-0.6cm}
	\caption{Summary statistics of the OULAD cohort used in our experiments.}
	\vspace{-0.3cm}
	\label{tab:oulad_stats}
	\begin{tabular}{lr}
		\hline
		\textbf{Quantity} & \textbf{Value} \\
		\hline
		Instances (module, presentation, student) & 32{,}593 \\
		Modules & 7 \\
		Presentations & 4 \\
		Module--presentation runs & 22 \\
		Successful ($y{=}1$) & 15{,}385 (47.2\%) \\
		At-risk (non-success, $y{=}0$) & 17{,}208 (52.8\%) \\
		\hline
	\end{tabular}
	\vspace{-0.8cm}
\end{table}

The target is a binary end-of-course outcome $y \in \{0,1\}$ derived from \textit{final\_result}. We set $y{=}1$ for successful outcomes (\textit{Pass} or \textit{Distinction}) and $y{=}0$ for unsuccessful outcomes (\textit{Fail} or \textit{Withdrawn}).
As shown in Table~\ref{tab:oulad_stats}, the benchmark includes $32{,}593$ instances from $22$ course runs (7 modules, 4 presentations), with $15{,}385$ successful (47.2\%) and $17{,}208$ unsuccessful (52.8\%) outcomes.

\subsection{Experimental Setup}
\label{sec:setup}

Timestamps in OULAD are expressed as day indices relative to course start and may be negative due to pre-course activity. We evaluate weekly cutoffs $\mathcal{T}=\{7,14,21,28,35,42,49,56\}$. For each $t \in \mathcal{T}$, we construct a cutoff-specific dataset $\mathcal{D}^{(t)}$ under the LEAP constraint, retaining only records available by day $t$ before any aggregation or transformation: \textit{studentVle.date} $\le t$ and, when assessment evidence is used, \textit{studentAssessment.date\_submitted} $\le t$. Leakage controls are enforced at each cutoff.

We distinguish interaction records, which capture engagement and practice, from assessment records, which reflect graded evaluation. From the filtered data, we compute a compact feature vector with deterministic handling of missingness (\(0\) if missing). Interaction features include total clicks, active days, unique resources, unique activity types, and daily click summary statistics. When assessment evidence is available, we also compute the number of submissions and the mean submission delay. Because OULAD does not report grade release times, score observability cannot be guaranteed from timestamps alone. We therefore adopt an explicit \emph{score-availability} assumption: if a submission occurs by cutoff $t$, its score is treated as observable at $t$. This is a dataset-specific choice rather than part of LEAP itself, which governs timestamp validity and truncation order.

In addition to the strict LEAP setting, we later report intentionally leaky variants as diagnostic controls. These variants are not alternative early-warning methods, but stress tests used to quantify how much performance can be inflated when temporal-validity constraints are violated.



\subsection{Benchmark Models}
\label{sec:models}

Since our goal is to evaluate temporal validity under leakage-controlled early availability, rather than to propose a new predictive architecture, we benchmark widely used classifiers from multiple model families implemented in scikit-learn \cite{pedregosa2011scikit}. For each cutoff $t$, models are trained independently on the corresponding dataset $\mathcal{D}^{(t)}$ to respect the information available at that time. To prevent preprocessing leakage and ensure consistent train--test transformations, feature scaling is applied only when required and is fit on the training split within the pipeline; tree-based ensembles are trained directly on the original feature scale.

\begin{table}[h]
	\centering
	\vspace{-0.7cm}
	\caption{Benchmark models and fixed configurations used in our experiments.}
	\vspace{-0.3cm}
	\label{tab:model_hparams}
	\begin{tabular}{l l l}
		\hline
		Models & Family & Hyper-parameter values \\
		\hline
		\LR & linear & standardized inputs; $\ell_2$-regularized logistic regression \\
		\RF & tree ensemble & 300 trees; bounded depth; minimum leaf size 2 \\
		\ET & tree ensemble & 300 trees; bounded depth; minimum leaf size 2 \\
		\GBDT & boosting (trees) & 250 stages; learning rate 0.05; depth 3 \\
		\AdaBoost & boosting & 200 stages; learning rate 0.5 \\
		\SVMLIN & kernel & standardized inputs; linear \SVM{} with probability outputs \\
		\SVMRBF & kernel & standardized inputs; RBF \SVM{} with probability outputs \\
		\KNN & instance-based & standardized inputs; $k{=}15$ neighbors \\
		\NB & Bayesian & Gaussian naive Bayes \\
		\MLP & shallow NN & standardized inputs; 2 hidden layers (64,32); 300 epochs \\
		\hline
	\end{tabular}
	\vspace{-0.8cm}
\end{table}

The benchmark set includes Logistic Regression (\LR), Random Forest (\RF), Extra Trees (\ET), Gradient Boosting (\GBDT), \AdaBoost, Support Vector Machines (\SVM), k-Nearest Neighbors (\KNN), Gaussian Naive Bayes (\NB), and a shallow Multi-Layer Perceptron (\MLP). To keep the comparison reproducible and computationally bounded, we use fixed hyperparameters without tuning; exact settings are reported in Table~\ref{tab:model_hparams}.

Evaluation uses an 80/20 stratified train/test split for each cutoff $t$. To reduce sensitivity to a single split, we repeat the procedure over five random seeds $s \in \{0,1,2,3,4\}$ and report mean and standard deviation across seeds. 


\subsection{Evaluation Metrics}
\label{sec:metrics}

All metrics are computed on the held-out test set using the predicted probability of the positive class ($y{=}1$). For a test set of size $n$, we denote the true labels by $y_i\in\{0,1\}$ and predicted probabilities by $\hat{p}_i\in[0,1]$.

\textit{Discrimination:} we report ROC-AUC, which evaluates how well the model ranks positive instances above negative ones across thresholds \cite{fawcett2006introduction}. We also report PR-AUC, computed as Average Precision (AP), which summarizes the precision--recall trade-off and is often more informative under class imbalance \cite{davis2006relationship,saito2015precision}.

\textit{Calibration:} we report the Brier score (lower is better), defined as the mean squared error between outcomes and predicted probabilities \cite{glenn1950verification}:

\vspace{-0.4cm}
\begin{equation}
	\mathrm{Brier} \;=\; \frac{1}{n}\sum_{i=1}^{n}(y_i-\hat{p}_i)^2.
	\vspace{-0.1cm}
\end{equation}
The Brier score is a strictly proper scoring rule for binary events \cite{gneiting2007strictly}.

\textit{Fixed-threshold performance: } we report the F1 score at threshold $0.5$ (F1@0.5), which balances precision and recall for the positive class \cite{van1979information}. For this metric, we convert probabilities to labels via $\hat{y}_i = 1$ if $\hat{p}_i \ge 0.5$ and $\hat{y}_i = 0$ otherwise:

\vspace{-0.3cm}
\begin{equation}
	\mathrm{F1} \;=\; \frac{2\,\mathrm{Prec}\,\mathrm{Rec}}{\mathrm{Prec}+\mathrm{Rec}},
	\qquad
	\mathrm{Prec}=\frac{\mathrm{TP}}{\mathrm{TP}+\mathrm{FP}},
	\quad
	\mathrm{Rec}=\frac{\mathrm{TP}}{\mathrm{TP}+\mathrm{FN}}.
	\vspace{-0.1cm}
\end{equation}
Having defined the leakage-excluded evaluation protocol, we next quantify its empirical implications and report performance across early observation windows.

\section{Experimental Results and Discussion}
\label{sec:results}

All results in this section are obtained under the strict LEAP protocol and are reported as mean$\pm$std over five random seeds. The analysis is organized according to the research questions introduced in Section~\ref{sec:introduction}.

\begin{figure}[h]
	\centering
	\vspace{-0.8cm}
	\includegraphics[width=0.49\linewidth]{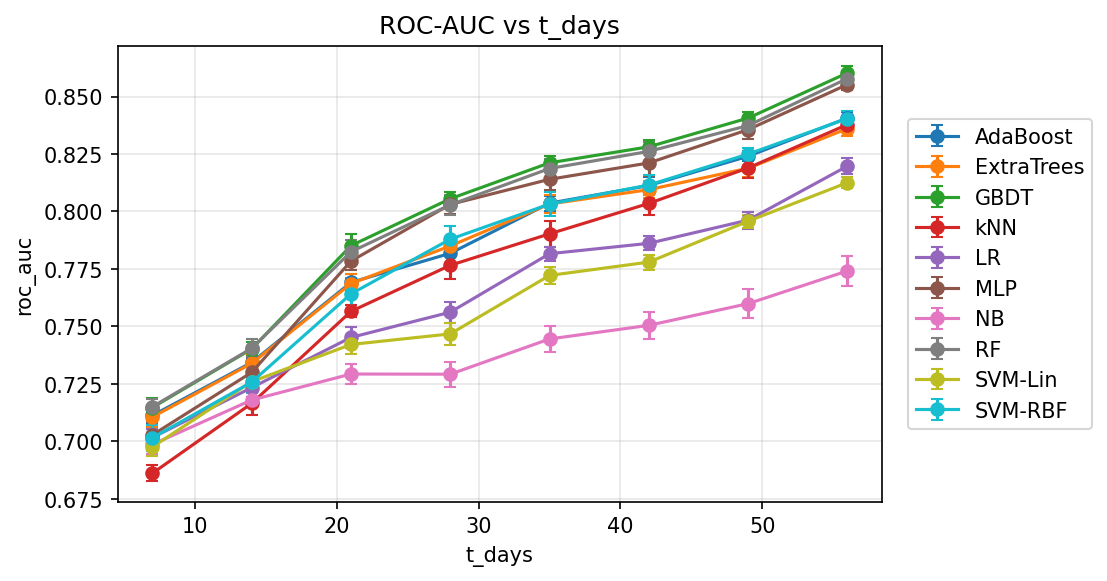}
	\includegraphics[width=0.49\linewidth]{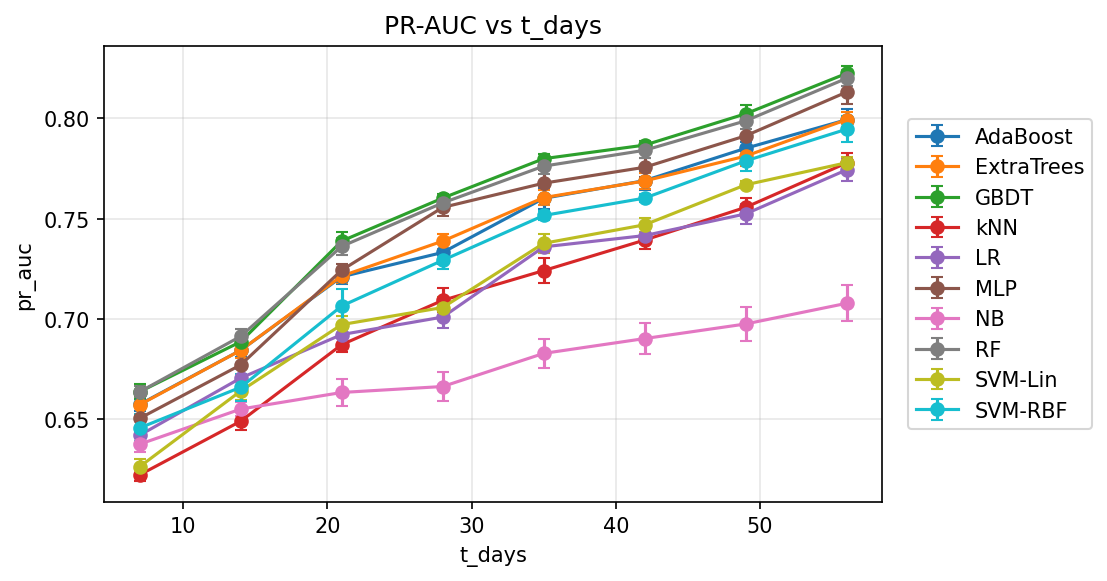}
	\vspace{-0.5cm}
	\caption{Earliness--performance curves under strict LEAP (mean$\pm$std over 5 seeds).}
	\vspace{-0.8cm}
	\label{fig:auc_vs_time}
\end{figure}

\subsection{Results}

\textit{RQ1 - Earliness--Performance Trends: }
Figure~\ref{fig:auc_vs_time} shows a clear and generally increasing earliness--performance relationship. Discrimination improves steadily as the observation window expands from $t{=}7$ to $t{=}56$ days. The best ROC-AUC increases from $0.7151\pm0.0035$ at $t{=}7$ to $0.8602\pm0.0028$ at $t{=}56$, while PR-AUC increases from $0.6638\pm0.0026$ to $0.8224\pm0.0034$. Calibration also improves with more evidence: the best Brier score decreases from $0.2132\pm0.0014$ at $t{=}7$ to $0.1511\pm0.0017$ at $t{=}56$. Standard deviations remain small across seeds, indicating that the observed trends are stable under repeated splits.

Performance gains are not uniform across time. The largest single improvement in best ROC-AUC occurs between $t{=}14$ and $t{=}21$ (from $0.7404$ to $0.7851$, i.e., $+0.0447$), suggesting that week 3 provides a substantial increase in predictive evidence beyond the first two weeks. This change point supports reporting results across multiple cutoffs rather than relying on a single early window.

\begin{table}[h]
	\centering
	\vspace{-0.7cm}
	\caption{Best ROC-AUC model per cutoff under strict LEAP (mean$\pm$std over 5 seeds).}
	\label{tab:best_per_cutoff}
	\vspace{-0.3cm}
	\begin{tabular}{c l c c c}
		\hline
		$t$ (days) & Model & ROC-AUC & PR-AUC & Brier \\
		\hline
		7  & \RF   & $0.7151\pm0.0035$ & $0.6638\pm0.0026$ & $0.2132\pm0.0014$ \\
		14 & \RF   & $0.7404\pm0.0039$ & $0.6914\pm0.0038$ & $0.2045\pm0.0013$ \\
		21 & \GBDT & $0.7851\pm0.0050$ & $0.7389\pm0.0044$ & $0.1878\pm0.0021$ \\
		28 & \GBDT & $0.8055\pm0.0030$ & $0.7603\pm0.0023$ & $0.1790\pm0.0014$ \\
		35 & \GBDT & $0.8212\pm0.0028$ & $0.7799\pm0.0022$ & $0.1717\pm0.0015$ \\
		42 & \GBDT & $0.8281\pm0.0029$ & $0.7867\pm0.0018$ & $0.1683\pm0.0015$ \\
		49 & \GBDT & $0.8406\pm0.0028$ & $0.8023\pm0.0042$ & $0.1620\pm0.0015$ \\
		56 & \GBDT & $0.8602\pm0.0028$ & $0.8224\pm0.0034$ & $0.1511\pm0.0017$ \\
		\hline
	\end{tabular}
	\vspace{-0.7cm}
\end{table}

\textit{RQ2 - Benchmark Model Rankings: }
Table~\ref{tab:best_per_cutoff} reports the best-performing benchmark model at each cutoff when selected by ROC-AUC. \RF{} is strongest in the earliest windows ($t{=}7$ and $t{=}14$), whereas \GBDT{} becomes dominant from $t{=}21$ onward and remains best through $t{=}56$. The ROC-AUC standard deviation is below $\approx 0.0068$ across cutoffs, suggesting that the ranking is not driven by a particular split.

\begin{table}[h]
	\centering
	\vspace{-0.7cm}
	\caption{All benchmark models at $t{=}56$ under strict LEAP (mean$\pm$std over 5 seeds).}
	\vspace{-0.3cm}
	\label{tab:all_models_t56}
	\begin{tabular}{l c c c c}
		\hline
		Model & ROC-AUC & PR-AUC & F1@0.5 & Brier \\
		\hline
		\GBDT       & $0.8602\pm0.0028$ & $0.8224\pm0.0034$ & $0.7777\pm0.0039$ & $0.1511\pm0.0017$ \\
		\RF         & $0.8578\pm0.0029$ & $0.8200\pm0.0039$ & $0.7781\pm0.0043$ & $0.1533\pm0.0017$ \\
		\MLP        & $0.8552\pm0.0024$ & $0.8131\pm0.0059$ & $0.7711\pm0.0071$ & $0.1548\pm0.0026$ \\
		\AdaBoost   & $0.8406\pm0.0025$ & $0.7994\pm0.0054$ & $0.7532\pm0.0030$ & $0.1891\pm0.0013$ \\
		\SVMRBF  & $0.8404\pm0.0033$ & $0.7946\pm0.0064$ & $0.7630\pm0.0042$ & $0.1612\pm0.0024$ \\
		\KNN        & $0.8378\pm0.0029$ & $0.7778\pm0.0047$ & $0.7652\pm0.0025$ & $0.1625\pm0.0016$ \\
		\ET & $0.8361\pm0.0033$ & $0.7994\pm0.0036$ & $0.7562\pm0.0053$ & $0.1698\pm0.0016$ \\
		\LR         & $0.8197\pm0.0035$ & $0.7744\pm0.0058$ & $0.7473\pm0.0043$ & $0.1748\pm0.0019$ \\
		\SVMLIN  & $0.8125\pm0.0024$ & $0.7779\pm0.0026$ & $0.7539\pm0.0025$ & $0.1759\pm0.0018$ \\
		\NB         & $0.7741\pm0.0065$ & $0.7079\pm0.0089$ & $0.7163\pm0.0074$ & $0.2375\pm0.0062$ \\
		\hline
	\end{tabular}
	\vspace{-0.7cm}
\end{table}

To compare model families under the largest information budget, Table~\ref{tab:all_models_t56} reports all benchmark models at $t{=}56$. \GBDT{} achieves the strongest overall performance (ROC-AUC $0.8602\pm0.0028$, PR-AUC $0.8224\pm0.0034$) and the best calibration (Brier $0.1511\pm0.0017$). \RF{} and \MLP{} are close runners-up, while \NB{} remains weaker in both discrimination (ROC-AUC $0.7741\pm0.0065$) and calibration (Brier $0.2375\pm0.0062$). The results indicate that classical tree-based ensembles provide robust performance throughout the earliness spectrum, with boosting offering consistent advantages once sufficient evidence accumulates.

\begin{figure}[h]
	\centering
	\vspace{-1.2cm}
	\includegraphics[width=0.99\linewidth]{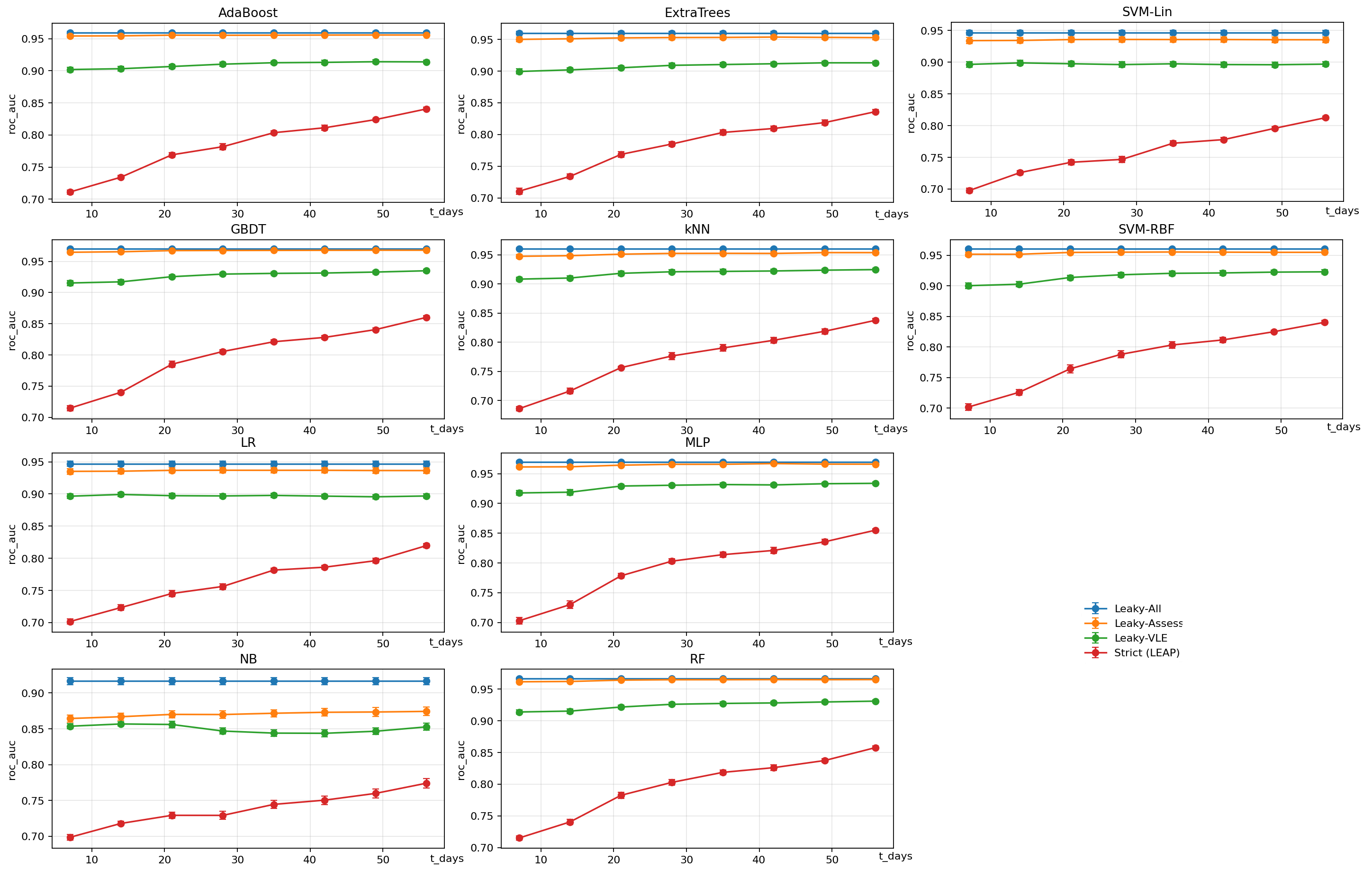}
	\vspace{-0.4cm}
	\caption{Leakage ablation: strict LEAP vs.\ intentionally leaky variants.}
	\vspace{-0.8cm}
	\label{fig:leakage_ablation}
\end{figure}

\textit{RQ3 - Leakage Ablation:}
Figure~\ref{fig:leakage_ablation} quantifies the impact of temporal leakage by comparing strict LEAP with intentionally leaky variants and reveals substantial inflation at early cutoffs. At $t{=}7$, \RF{} increases from ROC-AUC $0.7151$ (Strict) to $0.9669$ (Leaky-All), and \GBDT{} increases from $0.7147$ to $0.9704$. Most of this inflation is attributable to assessment leakage alone; for instance, \RF{} reaches ROC-AUC $0.9616$ under Leaky-Assessment. These results show that post-cutoff submissions and grades can make an ``early'' benchmark unrealistically optimistic and can invalidate early-warning conclusions. The ablation therefore provides empirical justification for the LEAP design: leakage control should be treated as an enforced benchmark requirement rather than an implicit assumption. Strict LEAP results constitute the valid early-prediction outcomes, while leaky variants serve only as diagnostic evidence of benchmark integrity.

\textit{RQ4 - Temporal Shift in Predictive Evidence: }
To characterize how predictive evidence evolves with earliness, we examine feature importance and linear coefficients across cutoffs. At early cutoffs, behavioral engagement dominates. At $t{=}7$, engagement volume and activity regularity are most influential; for example, \textit{total\_clicks\_t} is the top feature for \GBDT{} with importance $\approx 0.57$. As the observation window expands, the signal shifts toward observed assessment evidence. By $t{=}56$, assessment performance becomes the primary driver; \textit{avg\_score\_t} reaches importance $\approx 0.62$ for \GBDT{} and is also top-ranked for \RF{} (importance $\approx 0.33$). \LR{} coefficients corroborate this pattern, with early emphasis on engagement intensity and later dominance of assessment score.

Overall, these findings support the intended interpretation of the benchmark under strict early availability: the earliest predictions rely on behavioral proxies that are available prior to substantial graded evidence, while later cutoffs increasingly reflect assessment outcomes that are observable by the cutoff.

\subsection{Discussion}
\label{sec:discussion}

The results suggest that early-warning signals cannot be reduced to activity volume alone. In LMS settings such as OULAD, some learners may generate relatively few interaction traces while still obtaining good evaluation outcomes, for example because they study efficiently, rely on prior knowledge, or use external learning resources. Conversely, other learners may show intense platform activity but still perform poorly, which may indicate difficulty, repeated attempts, or ineffective learning strategies rather than successful progress. 

Prolonged absence of activity is often associated with disengagement, withdrawal, or failure, but it is not by itself a sufficient predictor, since some learners may re-engage later or complete required assessments with limited online traces. These patterns motivate distinguishing between \emph{practice} activity and \emph{evaluative} activity. Practice interactions (e.g., content views, forum consultation, or ungraded exercises) primarily reflect engagement and self-regulation, whereas evaluative records (e.g., graded quizzes and assessments) reflect performance on tasks defined and validated by the instructor as part of the pedagogical design.

This distinction is important because predictive gains at later cutoffs are likely driven not only by additional activity traces, but also by the increasing availability of instructor-mediated evaluation evidence. In OULAD, the timing of such evidence depends not only on when an assessment is submitted, but also on when the corresponding score becomes available to the learner, which is not explicitly recorded. For this reason, results involving assessment scores must be interpreted under the stated score-availability assumption. Beyond this limitation, future work should move from activity-only early warning toward richer pedagogical representations of the learner. Rather than relying exclusively on aggregate LMS traces, future systems could incorporate competency- and prerequisite-structured information, as well as ontology-grounded skill representations, to support more interpretable and pedagogically meaningful early-warning decisions~\cite{luyen2025automated,le2025well}.


\section{Conclusion}
\label{sec:conclusion}
This paper investigated early prediction of end-of-course outcomes from partially observed LMS interactions under strict temporal validity. Rather than proposing a new predictive architecture, it introduced LEAP as a protocol for leakage-controlled early-warning benchmarking. Experiments on OULAD across weekly cutoffs and multiple benchmark models showed a stable accuracy--earliness trade-off, with a notable performance gain around week~3; Random Forest performed best at the earliest cutoffs, while Gradient Boosting dominated at later cutoffs. Leakage ablations further demonstrated that post-cutoff information, especially assessment-related evidence, can severely inflate apparent ``early'' performance. 
Future work will extend LEAP to sequence-based models, more realistic evaluation splits, and integration into LMS platforms (e.g., Moodle). It will also explore context-aware early warning by incorporating factors beyond LMS traces, such as learner context, connectivity conditions, work environment, and personal circumstances, under ethical and privacy constraints.

\vspace{-0.5cm}
\begin{credits}
	\subsubsection{\ackname}
	We thank the Ikigai consortium led by the association Games for Citizens, the company Gamaizer, as well as the FORTEIM project (winner of the AMI CMA France 2030 call for projects), for their support and collaboration. Their contributions have provided significant added value to the completion of this research.
\end{credits}

	\bibliographystyle{splncs04}
	\bibliography{references}
	
\end{document}